\documentclass[10pt,twocolumn,letterpaper]{article}

\usepackage{iccv}
\makeatletter
\@namedef{ver@everyshi.sty}{}
\makeatother
\usepackage{tikz}
\usepackage{pgfplots}
\usetikzlibrary{shapes,arrows,decorations.pathreplacing}

\usepackage{times}
\usepackage{epsfig}
\usepackage{graphicx}
\usepackage{amsmath}
\usepackage{amssymb}

\usepackage{algorithm}
\usepackage{algorithmic}
\usepackage{microtype} 
\usepackage{subfigure}
\usepackage{amsfonts}
\usepackage{amsmath}
\usepackage{cancel}

\DeclareMathOperator*{\argmax}{arg\,max}
\DeclareMathOperator*{\argmin}{arg\,min}

\usepackage[latin1]{inputenc}

\usepackage[pagebackref=true,breaklinks=true,letterpaper=true,colorlinks,bookmarks=false]{hyperref}

\iccvfinalcopy 

\def\iccvPaperID{6575} 

\ificcvfinal\fi
\begin{document}

\title{Uncertainty Quantification in Deep Residual Neural Networks}

\author{Lukasz Wandzik \hspace{20pt} Raul Vicente Garcia \hspace{20pt} Jörg Krüger\\
Fraunhofer Institute for Production Systems and Design Technology\\
Pascalstraße 8-9, 10587 Berlin\\
{\tt\small \{Lukasz.Wandzik, Raul.Vicente, Joerg.Krueger\}@ipk.fraunhofer.de}
}

\maketitle

\begin{abstract}
Uncertainty quantification is an important and challenging problem in deep learning. Previous methods rely on dropout layers which are not present in modern deep architectures or batch normalization which is sensitive to batch sizes. In this work, we address the problem of uncertainty quantification in deep residual networks by using a regularization technique called stochastic depth. We show that training residual networks using stochastic depth can be interpreted as a variational approximation to the intractable posterior over the weights in Bayesian neural networks. We demonstrate that by sampling from a distribution of residual networks with varying depth and shared weights, meaningful uncertainty estimates can be obtained. Moreover, compared to the original formulation of residual networks, our method produces well-calibrated softmax probabilities with only minor changes to the network's structure. We evaluate our approach on popular computer vision datasets and measure the quality of uncertainty estimates. We also test the robustness to domain shift and show that our method is able to express higher predictive uncertainty on out-of-distribution samples. Finally, we demonstrate how the proposed approach could be used to obtain uncertainty estimates in facial verification applications.
\end{abstract}

\section{Introduction}
\label{sec:Introduction}
Uncertainty quantification plays an important role in many real-world computer vision applications, such as autonomous driving, cancer cell segmentation or facial recognition. Building safe and reliable vision systems requires algorithms capable of expressing uncertainty about their decisions.
However, most current deep learning based computer vision models \cite{Ledig17,SharpMask2016,Redmon2016,CycleGAN2017}  are unable to express uncertainty or extract reliable information about the confidence in their predictions. They provide point estimates, such as predictive probabilities from the \emph{softmax} layer, which are often misinterpreted as model confidence \cite{Gal2016}. 

\tikzset
{
	block/.style =  {draw, thick, rectangle, minimum height = 7em, minimum width = 0.5em}
}

\pgfmathdeclarefunction{gauss}{3}{%
	\pgfmathparse{1/(#3*sqrt(2*pi))*exp(-((#1-#2)^2)/(2*#3^2))}%
}

\pgfdeclarelayer{bg}    
\pgfsetlayers{bg,main}  

\begin{figure}[t]
\begin{center}
\begin{tikzpicture}[scale=0.8, every node/.style={scale=0.8}]
\node[inner sep=0pt] (in1) at (0,0) {\includegraphics[width=.1\textwidth]{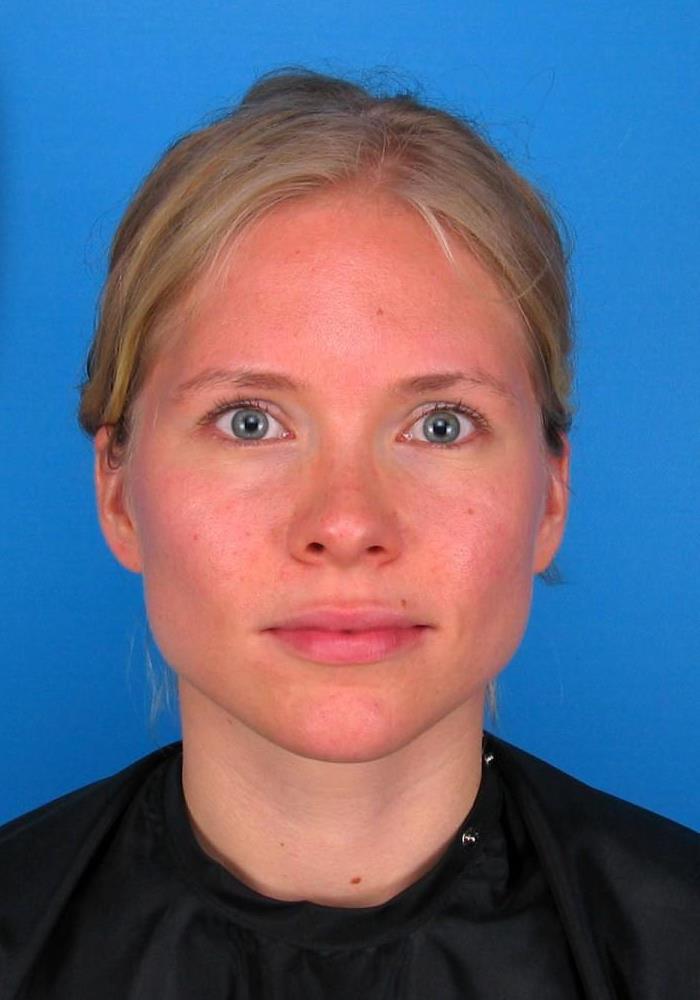}};
\node[inner sep=0pt] (in2) at (0.2,-0.2) {\includegraphics[width=.1\textwidth]{graphics/f4006.jpg}};
\node[inner sep=0pt] (in3) at (0.4,-0.4) {\includegraphics[width=.1\textwidth]{graphics/f4006.jpg}};

\node [block, draw, fill=white, label={\scriptsize CONV1}, right of=in3, xshift=1.5em] (conv1) {};

\node [block, draw=black!30!green, fill=black!30!green, label={\scriptsize R1}, right of=conv1, xshift=-0.75em] (r1) {};
\node [block, draw=red, fill=red, right of=r1, xshift=-1.75em] (r2) {};
\node [block, draw=black!30!green, fill=black!30!green, right of=r2, xshift=-1.75em] (r3) {};
\node [block, draw=black!30!green, fill=black!30!green, right of=r3, xshift=-1.75em] (r4) {};
\node [block, draw=black!30!green, fill=black!30!green, right of=r4, label={\small \ldots}, xshift=-1.75em] (r5) {};
\node [block, draw=red, fill=red, right of=r5, xshift=-1.75em] (r6) {};
\node [block, draw=black!30!green, fill=black!30!green, right of=r6, xshift=-1.75em] (r7) {};
\node [block, draw=black!30!green, fill=black!30!green, right of=r7, xshift=-1.75em] (r8) {};
\node [block, draw=red, fill=red, right of=r8, label={\scriptsize R9}, xshift=-1.75em] (r9) {};

\node [block, draw, fill=none, label={\scriptsize FC}, right of=r9, xshift=-0.75em] (fc) {};
\node [block, draw=none, fill=none, right of=fc, xshift=-0.75em] (dummy) {};

\draw [thick, decoration={brace, mirror, raise=3.25em}, decorate ] (r1.west) -- (r9.east) 
node [pos=0.5,anchor=north,yshift=-4.5em] {Residual blocks}; 
\path [->, thick, draw, -latex] (in3.east) -- node [midway] {} (conv1.west);
\begin{pgfonlayer}{bg} 
	\path [-, thick, draw] (conv1.east) -- node [midway] {} (r9.west);
\end{pgfonlayer}
\path [->, thick, draw, -latex] (r9.east) -- node [midway] {} (fc.west);

\path [->, thick, draw, -latex] (fc.east) -- node [midway] {} (dummy.west);

\path [-, thick, draw] ([xshift=-0.2em]r1.west) |- ([xshift=0.55em, yshift=-0.15em]r1.south) |- (r1.east);
\path [-, thick, draw] ([xshift=0.2em]r1.east) |- ([xshift=0.55em, yshift=-0.15em]r2.south) |- (r2.east);
\path [-, thick, draw] ([xshift=0.2em]r2.east) |- ([xshift=0.55em, yshift=-0.15em]r3.south) |- (r3.east);
\path [-, thick, draw] ([xshift=0.2em]r3.east) |- ([xshift=0.55em, yshift=-0.15em]r4.south) |- (r4.east);
\path [-, thick, draw] ([xshift=0.2em]r4.east) |- ([xshift=0.55em, yshift=-0.15em]r5.south) |- (r5.east);
\path [-, thick, draw] ([xshift=0.2em]r5.east) |- ([xshift=0.55em, yshift=-0.15em]r6.south) |- (r6.east);
\path [-, thick, draw] ([xshift=0.2em]r6.east) |- ([xshift=0.55em, yshift=-0.15em]r7.south) |- (r7.east);
\path [-, thick, draw] ([xshift=0.2em]r7.east) |- ([xshift=0.55em, yshift=-0.15em]r8.south) |- (r8.east);
\path [-, thick, draw] ([xshift=0.2em]r8.east) |- ([xshift=0.55em, yshift=-0.15em]r9.south) |- (r9.east);

\begin{axis}[
no markers, 
at={(0.825\linewidth,-0.15\linewidth)},
domain=0:6, 
samples=100,
ymin=0,
axis lines*=left, 
xlabel=$x$,
every axis y label/.style={at=(current axis.above origin),anchor=south},
every axis x label/.style={at=(current axis.right of origin),anchor=west},
height=3cm, 
width=4cm,
xtick=\empty, 
ytick=\empty,
enlargelimits=false, 
clip=false, 
axis on top,
grid = major,
hide y axis,
hide x axis
]

\addplot [very thick,cyan!50!black] {gauss(x, 3, 1)};

\pgfmathsetmacro\valueA{gauss(1,3,1)}
\pgfmathsetmacro\valueB{gauss(2,3,1)}

\draw [gray] (axis cs:3,0) -- (axis cs:3,0.4);
\draw [yshift=0.4cm, latex-latex](axis cs:1.5, 0) -- node [fill=white] {$\sigma^{2}$} (axis cs:4.5, 0);

\node[below] at (axis cs:3, 0)  {$\mu$}; 
\end{axis}

\end{tikzpicture}

\caption{Uncertainty quantification in deep residual neural networks. Green blocks depict active residual units whereas red blocks represent inactive units. Uncertainty estimates are obtained by performing several stochastic passes through the network while randomly switching $N$ residual blocks according to the Bernoulli distribution with probability $p_{i}$ for every block $i \in \left[1,\,N\right]$. The predictive variance $\sigma^{2}$ expresses how much the model is confident in its prediction. Image source: http://pics.stir.ac.uk}
\label{fig:pull_figure}
\end{center}
\end{figure}

A standard framework for reasoning about model uncertainty is given by the Bayesian probability theory. However, when applied to deep neural networks Bayesian inference becomes an intractable mathematical problem typically solved by means of approximation algorithms, such as Markov Chain Monte Carlo (MCMC) \cite{Bishop2006} or more recent Variational Inference (VI) \cite{Bishop2006}. Both methods exhibit slow convergence and low efficiency, especially with large datasets, which are inevitable in deep learning. Recently, the focus shifted to so-called \emph{stochastic regularization techniques} (SRT) such as dropout \cite{Srivastava2014} or batch normalization \cite{Ioffe2015} as efficient approximations to Bayesian neural networks. Gal and Ghahramani \cite{Gal2016PhD,Gal2016} proposed a method for estimating uncertainty in deep learning by using dropout layers. They showed that any neural network trained with dropout can be interpreted as a Bayesian approximation of a Gaussian process. Uncertainty estimates can be obtained by applying dropout masks at each forward pass and computing the variance or entropy of multiple predictions. However, most modern deep learning architectures, including residual networks \cite{He2016}, do not use dropout layers. Instead, they employ batch normalization, which itself is a strong regularization technique, rendering dropout ineffective when used within one model \cite{Huang2016}. Ioffe and Szegedy \cite{Ioffe2015} argue that batch normalization can be used instead of dropout while using both in one model creates redundancy and might increase training time.

\textbf{Contributions.} 
In this work, we study another regularization technique introduced by Huang \etal \cite{Huang2016}, called \emph{stochastic depth}. The approach exploits skip connections in deep residual networks by randomly dropping a subset of layers during training and utilizing the full depth of the network at test time. We build upon this idea and propose a novel approach for obtaining uncertainty estimates in deep residual networks. Unlike in Huang \etal \cite{Huang2016}, we do not utilize the full depth of the network at test time. Instead, we sample from a distribution of shorter networks and obtain the expected model output by averaging the predictions. This corresponds to performing several stochastic passes through the network while randomly dropping a subset of residual blocks. Each block is dropped independently with probability $p$ according to the Bernoulli distribution.
We show that our method is effective and provides meaningful uncertainty estimates for out-of-distribution samples. The approach does not require any additional layers and does not increases training time. Furthermore, it could be applied to any type of network with skip connections, including PyramidNet \cite{DHanKK17} and DenseNet \cite{HuangLMW17} or Dual Path Networks \cite{Chen17}. We present our findings on the CIFAR-10/100 and SVHN datasets and demonstrate an application of our method to face recognition.


\section{Related Work}
\label{sec:RelatedWork}
Most recent methods for model uncertainty estimation aim to approximate a Bayesian neural network \cite{Neal1996}, which is considered a gold standard in quantifying uncertainty.

\textbf{Backpropagation.}
Blundell \etal \cite{Blundell2015} proposed an algorithm called \emph{Bayes by Backprop}, in which the expected lower bound on the marginal likelihood is optimized in order to learn a probability distribution over the weights of a Bayesian neural network. Hern\'{a}ndez and Lobato \cite{Hernandez_Lobato2015} introduced a method for training Bayesian neural networks, called probabilistic backpropagation. They employ a collection of one-dimensional Gaussian distributions to approximate the marginal posterior distribution of individual weights. Both methods require modifications to the training algorithm and were only evaluated for simple models. 

\textbf{Optimization algorithms.}
Welling and Teh \cite{WellingT11} proposed an algorithm which approximates the MCMC sampling method by using stochastic gradient descent. They add Gaussian noise to each gradient update and adjust the noise level according to the learning rate. Khan \etal \cite{Khan2018} modified the Adam optimizer and obtained uncertainty estimates by perturbing the network's weights during gradient evaluations. Both methods require modifications to the respective optimization algorithm and are therefore limited to a particular optimizer.

\textbf{Deep Ensembles.}
Lakshminarayanan \etal \cite{Lakshminarayanan2017} trained ensembles of deep neural networks using adversarial examples. They treat the ensemble as an equally-weighted mixture model and combine the predictions by averaging the results. This technique bares a certain resemblance to our approach, however we assume that all the training data was generated by a single model or hypothesis instead of some linear combination of the models. Moreover, our model is not trained on adversarial examples and is more efficient due to extensive weight sharing among the particular sub-networks.

\textbf{Dropout.}
Gal and Ghahramani \cite{Gal2016, Gal2016PhD} proposed a method which aims to approximate the intractable true posterior by performing stochastic passes through a neural network. They show that dropout used at test time can be interpreted as variational inference approximation to the true posterior in Bayesian neural networks. However, modern deep architecture don't use dropout layers. Moreover, standard dropout becomes ineffective in conjunction with convolutional layers because of the strong spatial correlation of natural images. This method also has some similarities with our approach as it employs stochastic regularization techniques for uncertainty estimation.

\textbf{Batch normalization.} 
Teye \etal \cite{Teye18a} proposed a method for model uncertainty estimation by interpreting batch normalization as an approximate inference in Bayesian neural networks. The authors used mini-batch statistics to induce stochasticity in the test phase. They sample a batch from the training set and update the parameters in every batch normalization unit while performing several forward passes through the network. Atanov \etal \cite{Atanov2018} use a similar approach, they approximate the distribution of batch statistics with a fully-factorized parametric approximation. However, their method requires modifications to the batch normalization layers and is sensitive to batch sizes, which might be a limiting factor in some setting. 

\textbf{Multiplicative normalizing flows.} 
Louizos and Welling \cite{LouizosW17} used multiplicative noise as auxiliary random variables that augment the approximate posterior in a variational setting for Bayesian neural networks.

\section{Methods}
\label{sec:Methods}
In this section, we will show that training deep residual neural networks by randomly sampling a subset of residual blocks is equivalent to an approximate inference in Bayesian neural networks.
\subsection{Deterministic neural networks}
Given a set of observations $\mathcal{D} = \{(\mathbf{x}_{i}, \mathbf{y}_{i})\;|\;i \in \mathbb{N}\}$ and a model $f_{\mathbf{w}}$ with weights $\mathbf{w} \in \mathbb{R}$, the goal of training a deterministic neural network is to find optimal point estimates of model parameters that minimize a given cost function $\mathcal{L}$. This approach ignores the uncertainty in model selection and implies that the training data matches the parent population, which often leads to overconfident predictions.
 
The weights of a deterministic neural network can be found by maximum likelihood estimation (MLE)
\begin{equation}
\mathbf{w}^{\mbox{\tiny MLE}} = \argmax_{\mathbf{w}} \log p(\mathcal{D}|\mathbf{w})
\label{eq:det_mle}
\end{equation}
where $p(\mathcal{D}|\mathbf{w})$ is the data likelihood given the network's parameters and $\mathbf{w}^{\mbox{\tiny MLE}}$ is the point estimate. In order to prevent overfitting a prior over the weights of the network is often placed. The optimal parameters can be found by means of maximum a posteriori estimation (MAP)
\begin{align}
\begin{split}
	\mathbf{w}^{\mbox{\tiny MAP}} &= \argmax_{\mathbf{w}} \log p(\mathbf{w}|\mathcal{D})\\
	 &= \argmax_{\mathbf{w}} \log p(\mathcal{D}|\mathbf{w}) + \log p(\mathbf{w})
	 \label{eq:map}
\end{split}
\end{align}
where $\log p(\mathbf{w})$ acts as a regularizer or penalty term. In neural networks a $L_{2}$ regularization, also called weight decay is often used. It corresponds to placing a Gaussian prior over the weights and forcing their absolute value to be close to zero. 
\subsection{Bayesian neural networks}
In Bayesian neural networks (BNN) each weight is drawn from a prior distribution $p(\mathbf{w})$. Training a BNN consists of inferring the posterior distribution $p(\mathbf{w}|\mathcal{D})$ over the weights given the training data $\mathcal{D}$. The inference is performed according to the Bayes' theorem
\begin{equation}
p(\mathbf{w}|\mathcal{D}) = \frac{p(\mathcal{D}|\mathbf{w})p(\mathbf{w})}{p(\mathcal{D})}
\label{eg:bayes}
\end{equation}
The term $p(\mathcal{D})$ in the denominator is called model evidence and is intractable for most neural networks. The posterior predictive distribution of an unknown label $\mathbf{y}$ given an observation $\mathbf{x}$ can be calculated as follows
\begin{align}
\begin{split}
p(\mathbf{y}|\mathbf{x},\mathcal{D})&=\mathbb{E}_{p(\mathbf{w}|\mathcal{D})}[p(\mathbf{y}|\mathbf{x},\mathbf{w})]\\ &= \int\,p(\mathbf{y}|\mathbf{x},\mathbf{w})p(\mathbf{w}|\mathcal{D})\text{d}\mathbf{w}
\end{split}
\label{eq:pred_dist}
\end{align} 
Evaluating the above equation corresponds to employing an infinite number of neural networks and weighting their predictions according to the posterior distribution $p(\mathbf{w}|\mathcal{D})$. The predictive distribution of an unknown label $\mathbf{y}$ expresses the uncertainty of a model in its prediction.

\subsection{Variational approximation}
Since the integral of $p(\mathcal{D})$ is intractable and can not be evaluated analytically we have to resort to approximation algorithms.
Variational inference approximates the true posterior by an analytically tractable distribution $q$ parametrized by $\theta$. 
The parameters $\theta$ are optimized by minimizing the Kullback-Leibler (KL) divergence between the variational distribution $q(\mathbf{w}|\theta)$ and the true posterior $p(\mathbf{w}|\mathcal{D})$.
\begin{align}
\begin{split}
\theta^{\mbox{\tiny VI}}\!&=\argmin_{\mathbf{\theta}} \text{KL}\left[q(\mathbf{w}|\theta)||p(\mathbf{w}|\mathcal{D})\right] \\
\!&= \argmin_{\mathbf{\theta}} \text{KL}\left[q(\mathbf{w}|\theta)||p(\mathbf{w})\right] - \int\log p(\mathcal{D}|\mathbf{w})q(\theta)\mbox{d}\theta
\end{split}
\label{eq:vi}
\end{align}
The cost function in Eq. (\ref{eq:vi}) is often referred to as the evidence lower bound or ELBO. Minimizing the Kullback-Leibler divergence is equivalent to maximizing the ELBO. The integral in Eq. (\ref{eq:vi}) is intractable and cannot be evaluated analytically, therefore we approximate it using Monte Carlo integration over $\theta$ and arrive at the following objective
\begin{equation}
\mathcal{L}_{\mbox{\tiny VI}} := \text{KL}\left[q(\mathbf{w}|\theta)||p(\mathbf{w})\right] - \frac{1}{N}\sum_{n}^{N}\log p(\mathbf{y}_{n}|\mathbf{x}_{n},\mathbf{w}_{n})
\end{equation}
where $\mathbf{w}_{n}\sim q(\mathbf{w})$ and $N$ is the sample size. The term $\text{KL}\left[q(\mathbf{w}|\theta)||p(\mathbf{w})\right]$ can also be approximated following Gal and Ghahramani \cite{Gal2016PhD,Gal2016}, yielding the final objective 
\begin{equation}
\mathcal{L}_{\text{\tiny VI}} := \lambda \sum_{i\,=\,1}^{L} \lVert\mathbf{w}_{i}\rVert_{2}^{2} - \frac{1}{N}\sum_{n}^{N}\log p(\mathbf{y}_{n}|\mathbf{x}_{n},\mathbf{w}_{n})
\end{equation}
where $L$ is the number of layers. We can now replace the posterior distribution in Eq. (\ref{eq:pred_dist}) with the variational distribution $q(\mathbf{w})$ and approximate the integral with Monte Carlo integration following the derivation proposed by Gal and Ghahramani \cite{Gal2016}
\begin{equation}
	p(\mathbf{y}|\mathbf{x},\mathcal{D}) \overset{\mathrm{\scriptscriptstyle VI}}{\approx} \int p(\mathbf{x}|\mathbf{y},\mathbf{w})q(\mathbf{w})\text{d}\mathbf{w} \overset{\mathrm{\scriptscriptstyle MC}}{\approx}  \frac{1}{T} \sum_{t\,=\,1}^{T} p(\mathbf{y}|\mathbf{x}, \mathbf{w}_{t})
\end{equation}
where $\mathbf{w}_{t}\sim q(\mathbf{w})$ and $T$ denoting the number of stochastic forward passes through the neural network. 

\subsection{Deep residual networks}
Deep residual networks (ResNet) \cite{He2016} are modularized architectures consisting of stacked residual blocks organized in several stages (Fig. \ref{fig:pull_figure}). Residual networks employ skip connections, parallel to convolutional layers, which compute an identity function and make training of very deep networks possible. Each residual block performs the following computation	
\begin{equation}
	\mathbf{y}_{l} = \mathbf{x}_{l} + \mathcal{F}(\mathbf{x}_{l},\,\mathcal{W}_{l})
\end{equation}
where $\mathbf{x}_{l}$ and $\mathbf{y}_{l}$ are the input and output of the $l$-th residual block and $\mathcal{F}$ is the residual function. Furthermore, $\mathcal{W}_{l} = \{\mathbf{W}_{l,\,k} | 1 \leq k \le K \}$ is a set of weights associated with the $l$-th residual block, and $K$ is the number of layers in that block. 

\paragraph{Stochastic depth.} Huang \etal \cite{Huang2016} proposed a regularization technique for training of deep residual networks, called stochastic depth. The method exploits skip connection and randomly drops residual blocks replacing them with identity functions. For a given residual block with stochastic depth we get
\begin{equation}
\mathbf{y}_{l} = \mathbf{x}_{l} + \mathcal{F}(\mathbf{x}_{l}, \mathcal{W}_{i}) \odot b_{i}
\end{equation}
where $\odot$ denotes element-wise multiplication and $b_{i}$ is a random variable sampled from a Bernoulli distribution with probability $p_{i}$. This corresponds to drawing residual networks of varying depth according to the Binomial distribution.

\paragraph{Variational approximation in ResNets.}
We define the variational distribution $q(\mathbf{W}_{l,\,k})$ for every layer $k$ in a residual block $l$ as
\begin{equation}
\begin{split}
\mathbf{W}_{k} = \mathbf{M}_{k} \odot z_{k}\\
z_{k}\sim\mbox{Bernoulli}(p_{l})
\end{split}
\end{equation}
where $z_{k}$ is a Bernoulli distributed random variable with a probability $p_{l}$ for the $l$-th residual block, and $\mathbf{M}_{k}$ is a matrix of variational parameters. Following the derivation by Gal and Ghahramani \cite{Gal2016} we get the following objective
\begin{equation}
\mathcal{L}_{\text{MCSD}} := \lambda \sum_{i\,=\,1}^{L} \frac{p_{i}}{N}\lVert\mathbf{M}_{i}\rVert_{2}^{2} - \frac{1}{N}\sum_{n}^{N}\log p(\mathbf{y}_{n}|\mathbf{x}_{n},\mathbf{w}_{n})
\end{equation} 
where $\lambda$ is the weight decay, $p$ is the drop probability, $L$ is the number of layers in all residual stages and $N$ is the sample size.
We refer to our method as Monte Carlo Stochastic Depth or MCSD. The pseudocode for estimating predictive mean and entropy is summarized in Algorithm (\ref{alg:mcsd}). 

\subsection{Model uncertainty in face verification}
\label{sec:model_uncertainty}
In standard face verification, where a distance or score between pairs of feature vectors is computed, obtaining model or even predictive uncertainty is not possible. We propose a simple approach for obtaining model uncertainty estimates using the introduced method. Given two subject $A$ and $B$, we sample two sets of feature vectors by performing $T$ stochastic forward passes through the network. We compute the pairwise distances between set $A$ and $B$ using a given similarity metric $M$ and perform a verification for each subject given the optimal threshold $\tau$. Given the set of positive and negative decisions, we calculate the binary entropy
\begin{equation}	
\operatorname {\mathbb{H}\left[y\right]}=-y\log _{2}y-(1-y)\log _{2}(1-y)
\end{equation}
where $y$ denotes the verification decision. Similar to predictive entropy in classification the binary entropy in face verification can be interpreted as a measure of uncertainty.  

\begin{algorithm}[tb]
	\caption{MCSD}
	\label{alg:MCSD}
	\begin{algorithmic}
		\STATE {\bfseries Input:} Sample $\mathbf{x}$, number of stochastic passes $T$
		\STATE {\bfseries Output:} Mean prediction $\mathbf{\hat{y}}$, predictive entropy $\mathbb{H}[\mathbf{y}]$
		\FOR{$t \gets 1$ to $T$}
		\FOR{$l \gets 1$ to $L-1$}
		\STATE {Sample $b_{l}\,\sim\,\mbox{Bernoulli}(p_{l})$}
		\IF {$b_{i} \neq 0$}
		\STATE {$\mathcal{F}(\mathbf{x}_{l},\,\mathcal{W}_{l}) = \mathcal{F}(\mathbf{x}_{l},\,\mathcal{W}_{l})\,/\,(1\,-\,p_{l})$}
		\STATE {$\mathbf{x}_{l+1} = \mathbf{x}_{l} + \mathcal{F}(\mathbf{x}_{l},\,\mathcal{W}_{l})$}
		\ELSE
		\STATE {$\mathbf{x}_{l+1} = \mathbf{x}_{l}$}	
		\ENDIF 
		\ENDFOR
		\STATE {$\mathbf{y} = \mathbf{y} \cup \mathbf{x}_{L}$}
		\ENDFOR
		\STATE {$\mathbf{\hat{y}} = \mathbb{E}[\mathbf{y}]$}
		\STATE {$\mathbb{H}[\mathbf{y}] = -\sum_{c}\left[\frac{1}{T}\sum_{t}\mathbf{y}\log\left(\frac{1}{T}\sum_{t}\mathbf{y}\right)\right]$}
	\end{algorithmic}
\label{alg:mcsd}
\end{algorithm}

\section{Experiments}
\label{sec:Experiments}
We assess the uncertainty quality and probability calibration of our method on three publicly available datasets. Using standard evaluation metrics, we compare our approach to a deterministic model (baseline) and two popular methods based on spatial dropout \cite{TompsonGJLB15} and batch normalization \cite{Atanov2018}. We also perform experiments on out-of-distribution data and investigate the effectiveness of our method in a face verification scenario. 

\subsection{Evaluation metrics}
\textbf{Negative Log-likelihood (NLL)} is a standard measure of model quality proposed by Hastie \etal \cite{Hastie2001}. It expresses how well a given probabilistic model fits the underlying data. It is also commonly used to determine uncertainty quality. For a given probabilistic model $p$ and a test point $(\mathbf{y}_{i}, \mathbf{x}_{i})$, the NLL is defined as
\begin{equation}
\text{NLL}(\mathbf{y}_{i}, \mathbf{x}_{i}) = -\sum_{i\,=\,1}^{N}\log(p(\mathbf{y}_{i}|\mathbf{x}_{i}))
\end{equation}
whereas an approximation in our setting yields
\begin{equation}
\text{NLL} \approx -\log \frac{1}{N}\sum_{i\,=\,1}^{N}\frac{1}{T}\sum_{t\,=\,1}^{T}p(\mathbf{y}_{i}|f_{\mathbf{{w}}_{t}}(\mathbf{x}_{i}))
\end{equation}

with $T$ denoting the number of stochastic passes, $f_{\mathbf{w}}$ denotes the evaluated model with weights $\mathbf{w}$, and $N$ is the sample size. Since the quantity is negative, lower values indicate better model fit and higher uncertainty quality.
\paragraph{Brier Score (BS)} measures the accuracy of probabilistic predictions. It is defined as the mean squared deviation from empirical probabilities \cite{GuoPSW17}. For a given set of predictions lower score values indicate better calibrated predictive probabilities.  
\begin{equation}
\mbox{BS} = \frac{1}{N} \sum_{i\,=\,1}^{N}(f_{\mathbf{w}}(\mathbf{x}_{i}) - \mathbf{y}_{i})^{2}
\end{equation}
\paragraph{Expected Calibration Error (ECE)} was introduced by Naeini \etal \cite{Naeini2015} and is a measure of miscalibration. It is defined as the difference in expectation between confidence and accuracy. The ECE measure is calculated by partitioning predictions into M equally-spaced bins and taking a weighted average of absolute differences between accuracy and confidence in each bin $\text{B}_{m}$.
\begin{equation}
\mbox{ECE} = \sum_{m\,=\,1}^{M} \frac{|\mbox{B}_{m}|}{N} \bigg|\mbox{Acc}(|\mbox{B}_{m}) - \mbox{Conf}(|\mbox{B}_{m})\bigg|
\end{equation}

\subsection{Datasets}
\paragraph{CIFAR-10/100 and SVHN.}
We evaluate our approach on three popular and publicly available datasets. The CIFAR-10 \cite{Krizhevsky09} consists of 60k color images divided into 10 classes, with 6k images per class. The CIFAR-100 dataset is similar to CIFAR-10, except that it has 100 classes, each containing 600 images. We evaluate our approach on both datasets due to their similar image content but different target distribution. In contrast to many previous works we do not use the MNIST dataset for evaluation. Instead, we employ images from the SVHN dataset \cite{Netzer2011}, which comes from a significantly harder, real-world problem. Compared to MNIST, the SVHN dataset is much larger and contains over 600k images of house numbers distributed over 10 classes. 

\paragraph{VGGFace2.}
The VGGFace2 \cite{Cao18} is a large-scale image dataset for face recognition. It contains about 3.3 million face images of over 8k subjects with large variations in pose, age, illumination and ethnicity. We select the first 1000 subjects for training the face verification model, which amounts to a total of about 350k images. We also select a smaller subset of the first 100 subjects and train a second model for comparison. We refer to the larger subset with 1000 subjects as VGGFace2-1k and to the second subset with 100 subjects as VGGFace2-100. Furthermore, we use images from the Utrecht ECVP face dataset to select the optimal threshold for the face verification task. 
\paragraph{MultiPIE.}
We use the Multi-PIE \cite{Gross2010} dataset as a basis for the preparation of facial morphs. The original dataset contains images of 337 subjects from four different sessions. We only consider images with a neutral pose, facial expression and frontal illumination. We automatically generate facial morphs with varying blending factor, ranging form $0.1$ to $0.9$, using the splicing-morph approach. The exact procedure was described in \cite{Wandzik2017}. 

\subsection{Evaluation procedure}
We train and compare the four following methods using the introduced evaluation metrics: 
\begin{itemize}
	\item We refer to the deterministic residual network in its original formulation as DET, and use it as our baseline. 
	\item We refer to the method proposed by Gal and Ghahramani \cite{Gal2016}, called Monte Carlo Dropout as MCDO. In contrast to the original work, we do not use standard dropout. Instead, we follow the recommendations given by Li \etal \cite{Li2018} and Tompson \etal \cite{TompsonGJLB15} and use spatial dropout layers after each batch normalization unit. Tompson \etal \cite{TompsonGJLB15} reported that standard dropout is ineffective in fully convolutional networks due to strong spatial correlation of natural images. Consequently, the feature map activations are also strongly correlated and dropping them out independently has a negligible effect on the network. 
	\item We refer to the method proposed by Atanov \etal \cite{Atanov2018}, called Stochastic Batch Normalization as SBN. This method has no additional parameter apart from the sampling strategy.
	\item We refer to our own method described in Section \ref{sec:Methods} as MCSD.
\end{itemize}

We evaluate all methods using the ResNet-110 architecture with 110 layers and 54 residual blocks. We use the Type-A skip connection \cite{He2016} for identity mapping with no additional weights. All models are trained using batch normalization with shift and scale parameters and a batch size of 32. We employ a linear decay policy for the drop rates of residual units throughout our experiments as proposed by Huang \etal \cite{Huang2016} . The decision is motivated by the observation made by Greff \etal \cite{GreffSS16}. The authors argue that \emph{each consecutive residual block learns an iteratively refined estimation of data representation}. To make the comparison fair, we also use a linear decay policy for the drop rates in the MCDO models and optimize the drop probabilities using linear search. 

We evaluate the robustnesses of our approach to domain shift and out-of-distribution samples by training all models on the CIFAR-10 dataset and using the SVHN data for testing. We repeat the experiment by doing the opposite and evaluate the SVHN model on CIFAR-10 test data. In order to qualitatively compare the results of all four methods we use the CDF/Entropy plots proposed by Louizos and Welling \cite{LouizosW17}.

For the face verification experiments, we train two models on the VGGFace2-100 and VGGFace2-1k dataset respectively. Both models use the ResNet-152 architecture with bottleneck residual blocks and Type-A skip connections. The VGGFace2-100 model is trained with less data in order to show that our method provides meaningful epistemic uncertainty estimates that can be explained away with more data. We train both models using the softmax and cross entropy loss. For testing, we remove the last fully connected layer and normalize the output. In order to verify the identity of each individual subject we employ the cosine similarity metric.

\subsection{Morphing attack}
\begin{figure}[t]
	\begin{center}
		\centerline{
		\includegraphics[width=0.3\linewidth]{graphics/f4006.jpg}
		\includegraphics[width=0.3\linewidth]{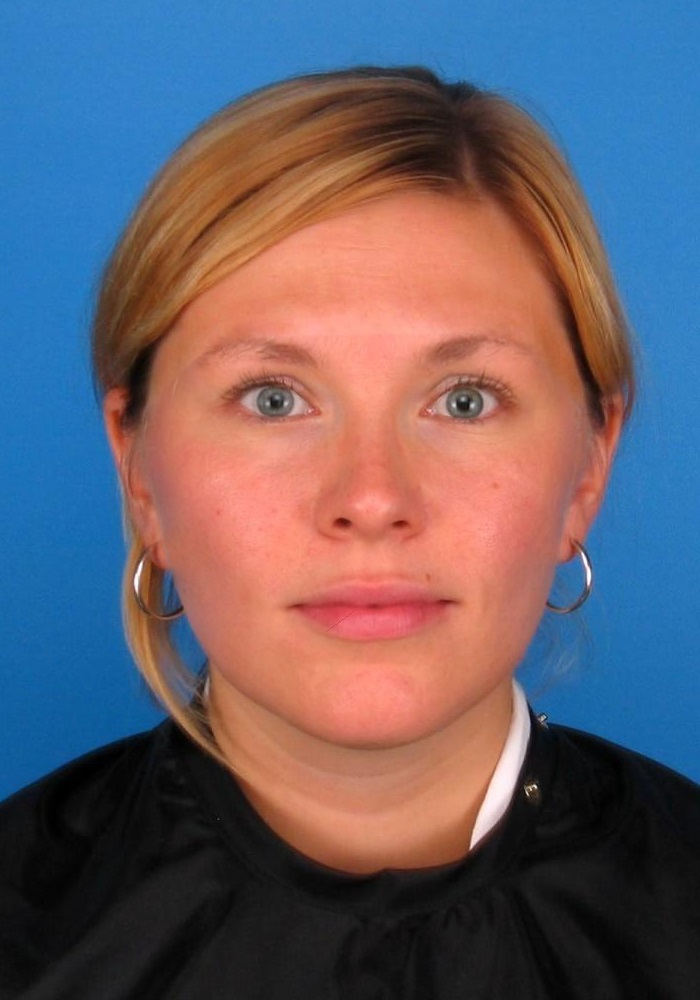}
		\includegraphics[width=0.3\linewidth]{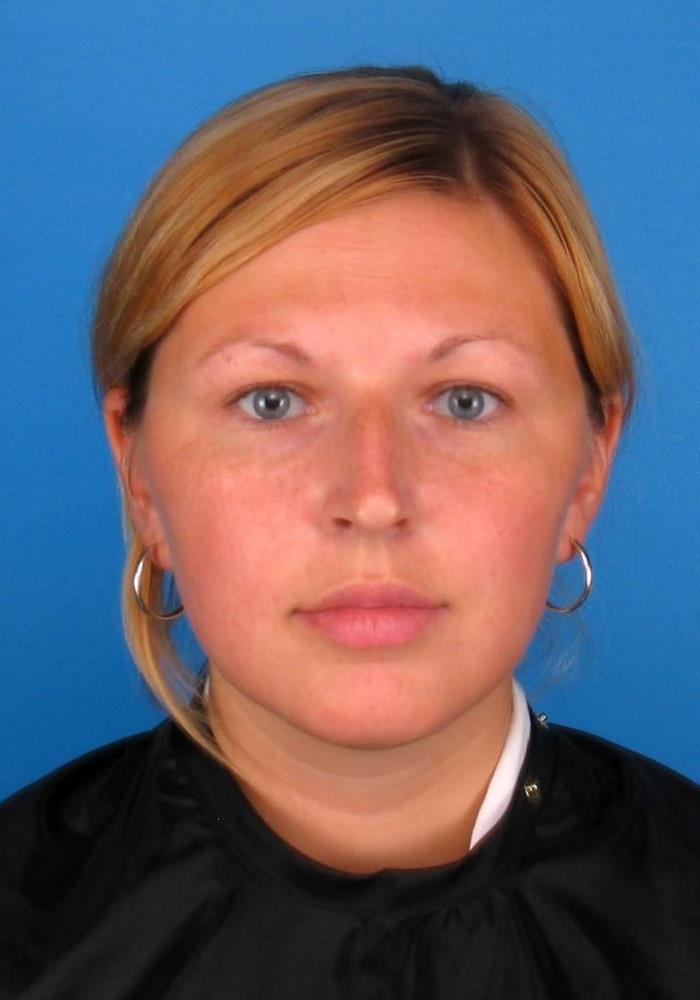}
		}
	\end{center}
	\caption{Example of a splicing morph (middle). Left: impostor image (source), middle: morphed image (template), right: accomplice image (target). Images were registered and morphed with a blending factor of $0.5$. 
	Poisson blending was used to adjust the color space of the source image. Image source: http://pics.stir.ac.uk}
	\label{fig:morphing_attack}
\end{figure}

In this experiment we test the vulnerability of our method to morphing attacks. Ferrara \etal \cite{Ferrara2014} show that facial recognition systems can be deceived by tampering the face image template. Morphing attacks are typically performed by blending two or more registered face images into a single image, which exhibits biometric characteristics of all involved persons. Formally, for a given distance metric $\mathbf{M}$ a successful morphing attack is defined as
\begin{equation}
\|I_{T} - I_{A}\|_{\mathbf{M}} < \tau \; \land \; \|I_{T} - I_{I}\|_{\mathbf{M}} < \tau
\end{equation}
where $I_{T}$ is the template, $I_{A}$ is the accomplice image, $I_{I}$ is the impostor image and $\tau$ is the distance threshold used for decision making (see Figure \ref{fig:morphing_attack}). Morphing attacks are easy to perform, even by non-experts and might therefore pose a serious threads in critical applications of face recognition like automatic border control.

\section{Results}
\label{sec:Results}
\subsection{Uncertainty quality}
In Table \ref{tab:table_cifar10} we present the results on the CIFAR-10 dataset. Our method outperforms other approaches in uncertainty quality and probability calibration. The test error is slightly higher than in the DET model, but the difference is negligible. The SBN method yields the worst results with very high NLL, indicating poor uncertainty quality.
\begin{table}[t]
	\caption{Evaluation results for uncertainty quality and probability calibration on the CIFAR-10 dataset. 
		    Results average 50 stochastic forward passes. Smaller values are better.}
	\label{tab:table_cifar10}
	\begin{center}
		\begin{tabular}{|l|c|c|c|c|}
			\hline
									        & \textbf{Test error} & \textbf{BS} 		  & \textbf{ECE} 	  & \textbf{NLL} \\
			\hline\hline
			\textbf{DET} \cite{He2016} 		& $\mathbf{0.056}$    & $0.045$ 	 		  & $0.199$           & $0.270$ \\ 
			\hline
			\textbf{MCDO} \cite{Gal2016} 	& $0.072$ 			  & $\mathbf{0.037}$ 	  & $0.041$  	      & $0.200$ \\
			\hline
			\textbf{SBN} \cite{Atanov2018} 	& $0.085$			  & $0.062$   		  	  & $0.115$  		  & $0.358$ \\	
			\hline
			\textbf{MCSD (ours)}			& $0.058$ 		      & $0.039$      & $\mathbf{0.041}$  & $\mathbf{0.189}$ \\
			\hline
		\end{tabular}
	\end{center}
\end{table}
Table \ref{tab:table_cifar100} shows the results on the CIFAR-100 dataset. Here the MCDO method consistently outperforms all other approaches in uncertainty quality, probability calibration and test error. However, our approach outperforms the baseline and yields comparable results to MCDO with a marginal difference in uncertainty quality and probability calibration. The SBN method improves upon the baseline only in calibration error.
\begin{table}[t]
	\caption{Evaluation results for uncertainty quality and probability calibration on the CIFAR-100 dataset. 
		Results after 50 stochastic forward passes. Smaller values are better.}
	\label{tab:table_cifar100}
	\vskip 0.15in
	\begin{center}
		\begin{tabular}{|l|c|c|c|c|}
			\hline
											& \textbf{Test error} & \textbf{BS} 		  & \textbf{ECE} 	  & \textbf{NLL} \\
			\hline\hline
			\textbf{DET} \cite{He2016} 		& $0.261$    		  & $0.169$ 	 		  & $0.191$           & $1.355$ \\ 
			\hline
			\textbf{MCDO} \cite{Gal2016} 	& $\mathbf{0.243}$ 	  & $\mathbf{0.122}$ 	  & $\mathbf{0.027}$   & $\mathbf{0.901}$ \\
			\hline
			\textbf{SBN} \cite{Atanov2018} 	& $0.305$			  & $0.175$   		  	  & $0.160$  		  & $1.472$ \\	
			\hline
			\textbf{MCSD (ours)}			& $0.255$ 	  		  & $0.126$      		  & $0.032$  		  & $0.936$ \\
			\hline
		\end{tabular}
	\end{center}
	\vskip -0.1in
\end{table}
In Table \ref{tab:table_svhn} we show the results on the SVHN dataset. MCDO outperforms all other approaches in test error, probability calibration and uncertainty quality. However, the results provided by MCDO and MCSD are very similar and outperform the baseline only marginally. This might be due to the strong regularizing effect of large datasets such as the SVHN. The inherent noise present in the data acts as a regularizer and diminishes the impact of dropout and similar techniques. This is not only a problem of our approach but any regularization technique, as can be seen in the results.
\begin{table}[t]
	\caption{Evaluation results for uncertainty quality and probability calibration on the SVHN dataset. 
		Results after 50 stochastic forward passes. Smaller values are better.}
	\label{tab:table_svhn}
	\vskip 0.15in
	\begin{center}
		\begin{tabular}{|l|c|c|c|c|}
			\hline
											& \textbf{Test error} & \textbf{BS} 		  & \textbf{ECE} 	  & \textbf{NLL} \\
			\hline\hline
			\textbf{DET} \cite{He2016} 		& $0.034$    		  & $0.024$  		      & $\mathbf{0.033}$  & $0.132$ \\ 
			\hline
			\textbf{MCDO} \cite{Gal2016} 	& $\mathbf{0.031}$ 	  & $\mathbf{0.023}$ 	  & $0.061$  	      & $\mathbf{0.121}$ \\
			\hline
			\textbf{SBN} \cite{Atanov2018} 	& $0.111$			  & $0.059$   		  	  & $0.066$  		  & $0.401$ \\	
			\hline
			\textbf{MCSD (ours)}			& $0.033$ 	  		  & $0.024$      		 & $0.050$  		  & $0.128$ \\
			\hline
		\end{tabular}
	\end{center}
	\vskip -0.1in
\end{table}

\subsection{Probability calibration}
\begin{figure}[ht]
	\vskip 0.2in
	\begin{center}
		\centerline{\includegraphics[width=1.0\columnwidth]{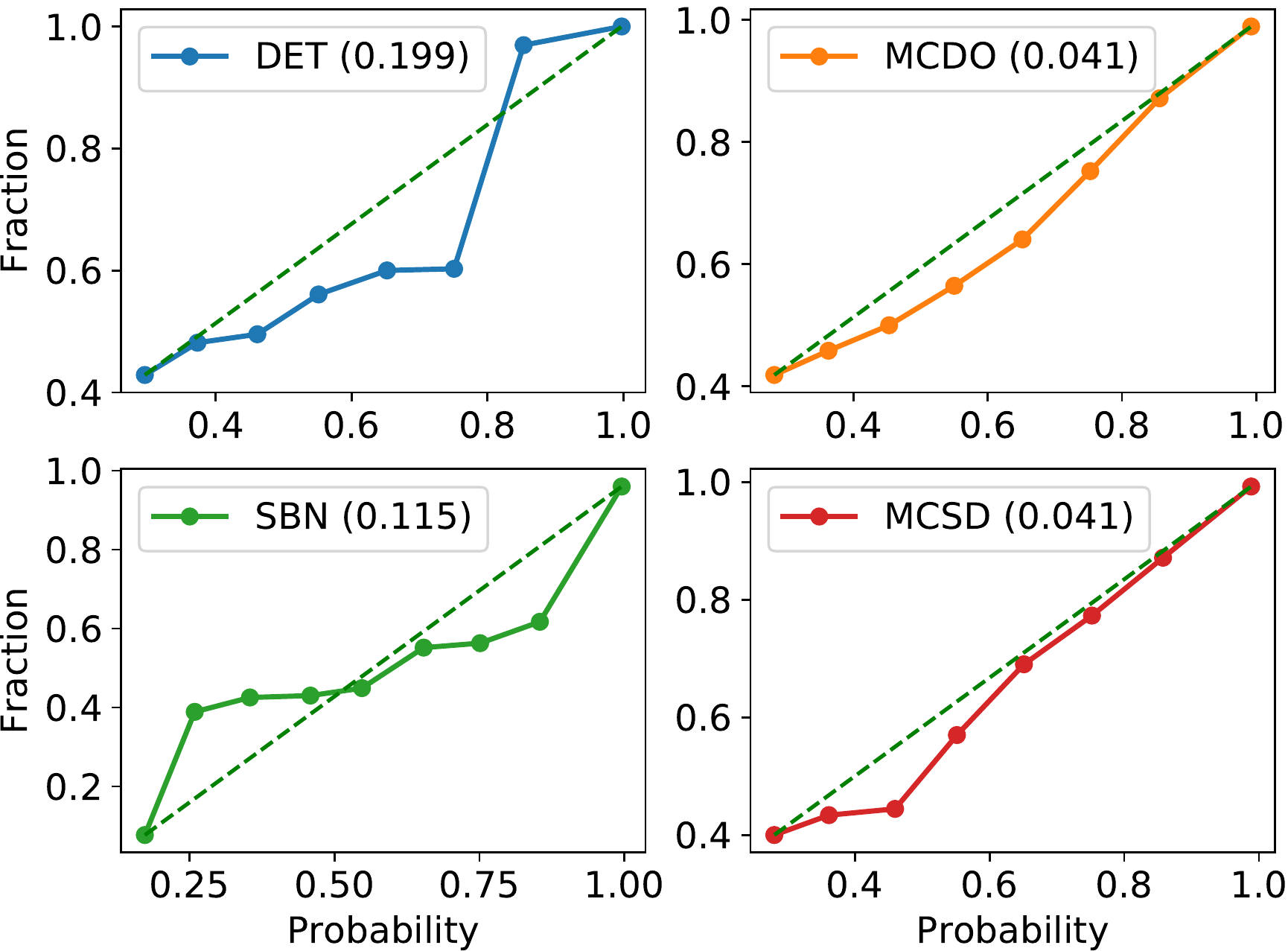}}
		\caption{Reliability diagrams for the CIFAR-10 dataset. A perfect calibration is depicted by the dashed line. Readings above the dashed line indicate overconfident predictions. In brackets we report the ECE for each method.}
		\label{fig:calibration_cifar10}
	\end{center}
	\vskip -0.2in
\end{figure}

\begin{figure}[ht]
	\vskip 0.2in
	\begin{center}
		\centerline{\includegraphics[width=1.0\columnwidth]{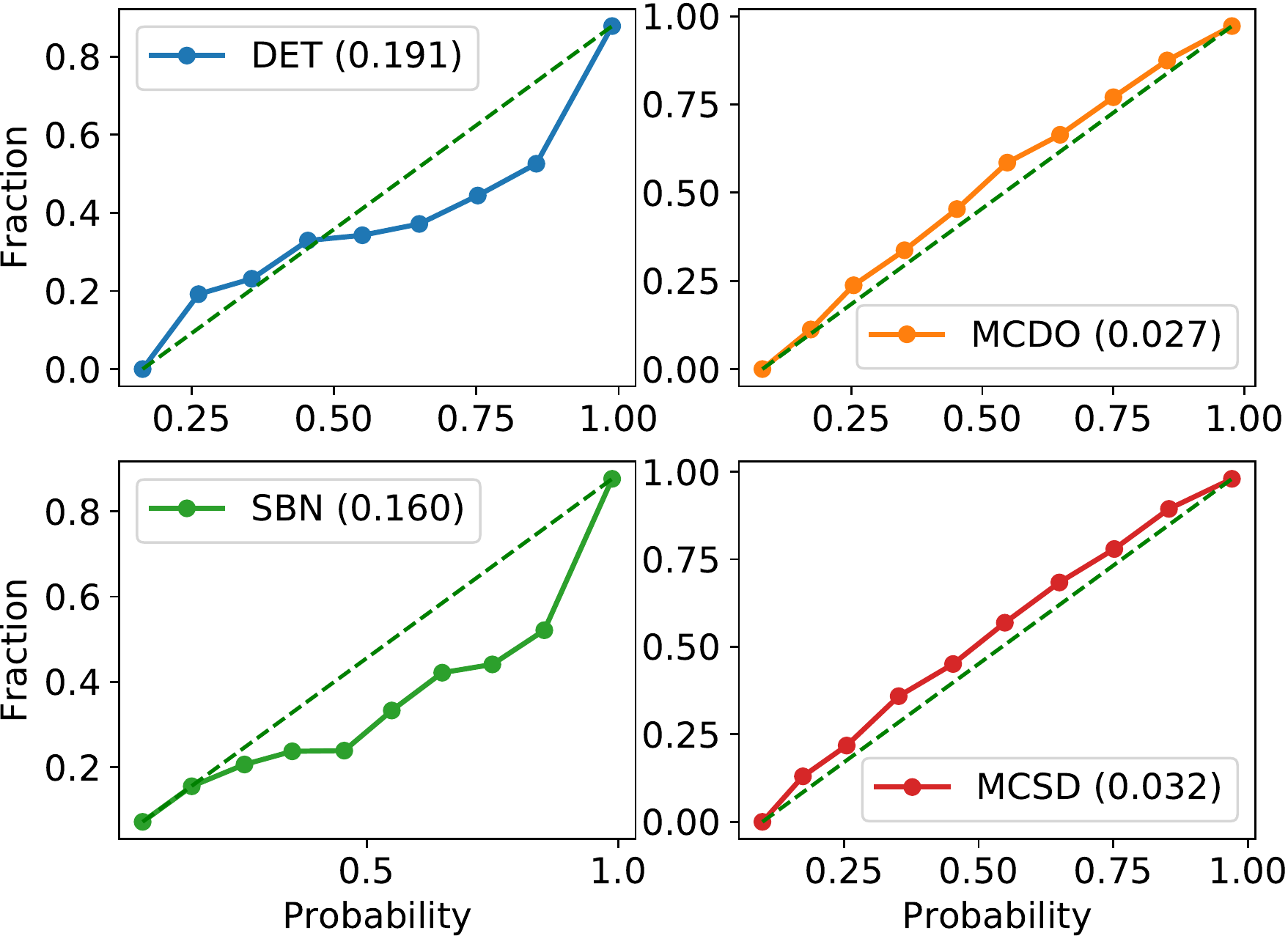}}
		\caption{Reliability diagrams for the CIFAR-100 dataset. A perfect calibration is depicted by the dashed line. Readings above the dashed line indicate overconfident predictions. In brackets we report the ECE for each method.}
		\label{fig:calibration_cifar100}
	\end{center}
	\vskip -0.2in
\end{figure}

\begin{figure}[ht]
	\vskip 0.2in
	\begin{center}
		\centerline{\includegraphics[width=1.0\columnwidth]{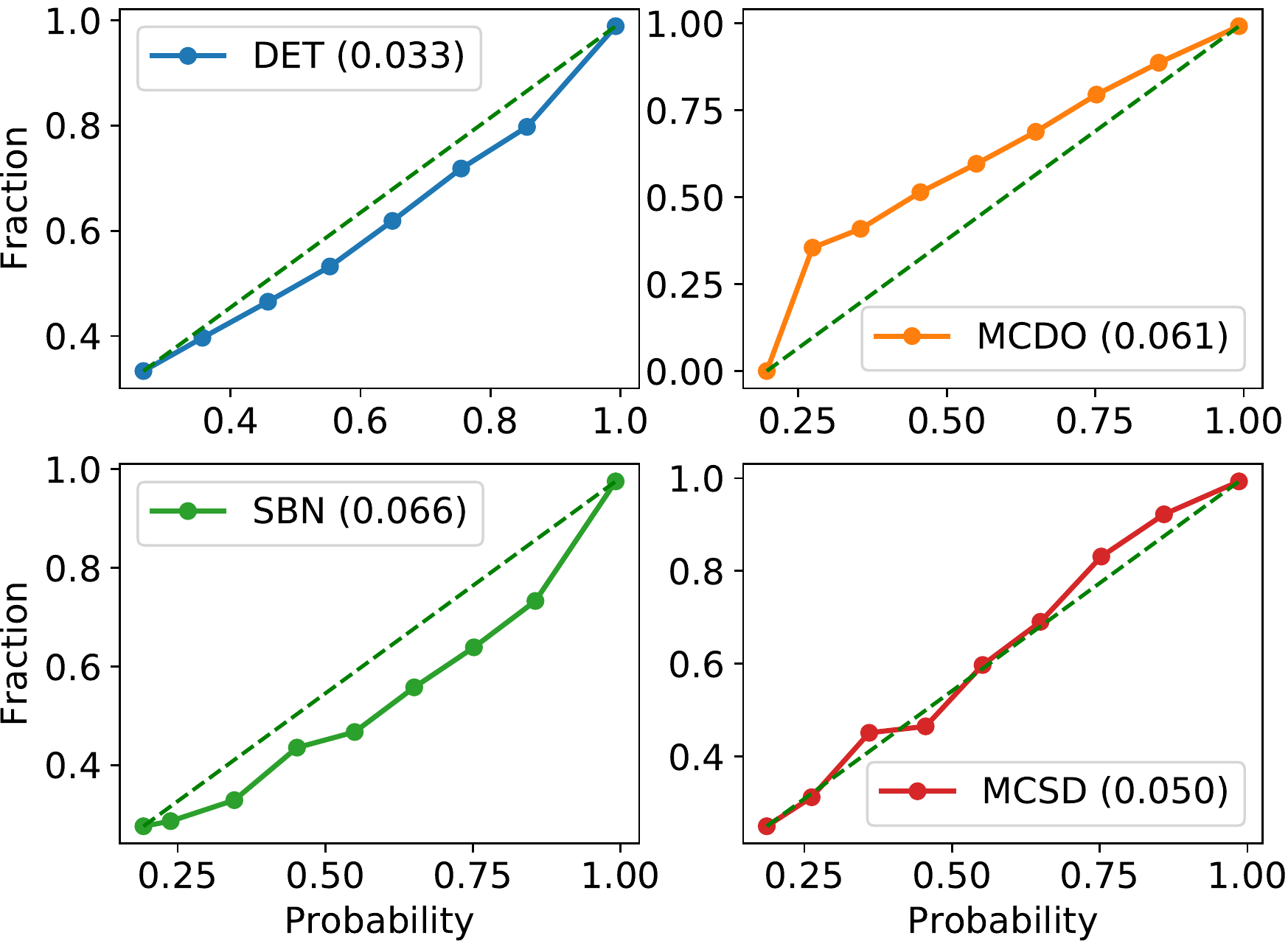}}
		\caption{Reliability diagram for the SVHN. A perfect calibration is depicted by the dashed line. Readings above the dashed line indicate overconfident predictions. In brackets we report the ECE for each method.}
		\label{fig:calibration_svhn}
	\end{center}
	\vskip -0.2in
\end{figure}

We present qualitative results for probability calibration using reliability diagrams. In Figure \ref{fig:calibration_cifar10}, \ref{fig:calibration_cifar100} and \ref{fig:calibration_svhn} the results for all four methods are shown. Reliability diagrams plot expected sample accuracy as a function of confidence. A perfect calibration is represented by the identity function. Our method outperforms the baseline and SBN consistently on all datasets and yields comparable results to MCDO. The DET and SBN approaches provide underconfident predictions on the CIFAR-100 (\ref{fig:calibration_cifar100}) and SVHN (\ref{fig:calibration_svhn}) datasets. MCDO is underconfident on the CIFAR-10 (\ref{fig:calibration_cifar10}) and overconfident on the SVHN (\ref{fig:calibration_svhn}) dataset. Our approach is the most stable of all methods, deviating only marginally form perfect calibration on all three datasets.  

\subsection{Domain shift}
\begin{figure}[t]
	\begin{center}
		\centerline{\includegraphics[width=1.0\linewidth]{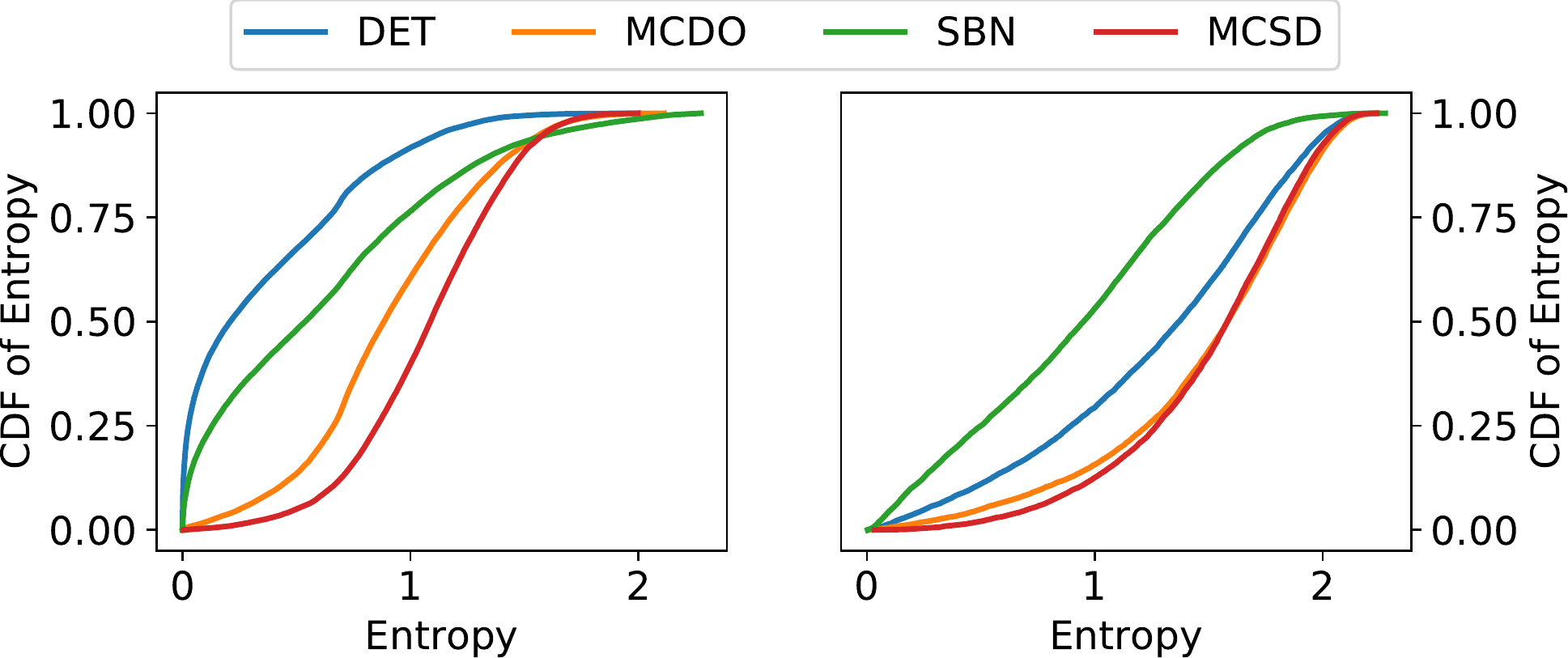}}
	\end{center}
	\caption{Empirical CDF of entropy for out-of-distribution data. Curves closer to the bottom right corner represent better calibrated models and indicate low probability of high confidence predictions. Results using SVHN (left) and CIFAR-10 (right) as test data.}
	\label{fig:domain_shift_cifar_svhn}
\end{figure}

We present the results on domain shift in Figure \ref{fig:domain_shift_cifar_svhn}. Our method outperforms all other approaches in this experiment being the most sensitive to out-of-distribution data.
The baseline method is the most overconfident on out-of-distribution samples when trained on the CIFAR-10 dataset. The picture changes for the SVHN dataset. All methods are much less confident when using CIFAR-10 as test data. This is probably due to the strong regularizing effect of this large dataset which prevents over-fitting.

\subsection{Morphing attack}

As mentioned in the previous section we evaluate two models trained on the VGGFace2-100 and VGGFace2-1k datasets respectively. We select the optimal threshold at $0.1 \%$ false acceptance rate, which is the prescribed value for face recognition systems in border control scenarios. We increase the blending factor and measure the accuracy and predictive entropy using the proposed method.

\begin{figure}[t]
	\begin{center}
		\centerline{\includegraphics[width=1.0\linewidth]{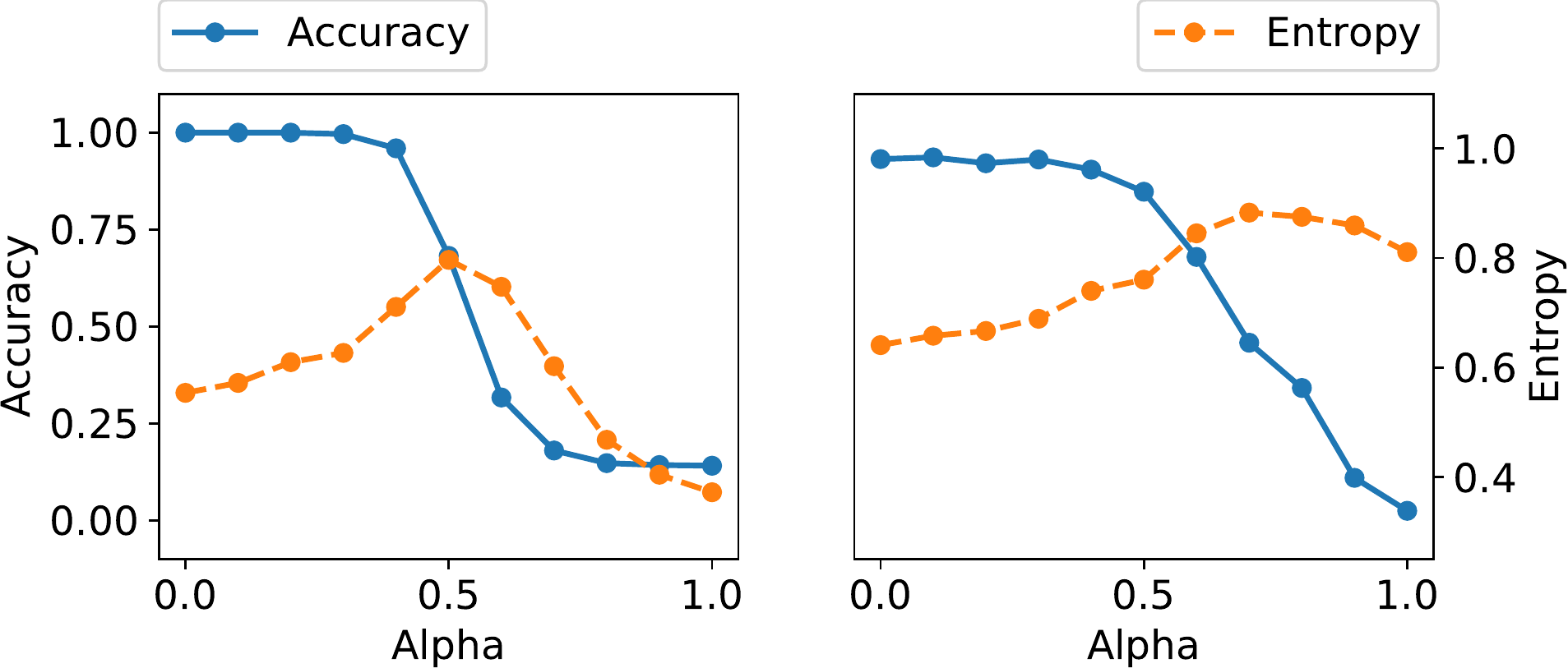}}
	\end{center}
	\caption{Accuracy/entropy for a varying blending factor alpha. Results after 50 stochastic forward passes for the VGGFace2-1k (left) and VGGFace2-100 (right) trained models.}
	\label{fig:entropy_morphing_attack}
\end{figure}

\begin{figure}[ht]
	\begin{center}
		\centerline{\includegraphics[width=0.5\columnwidth]{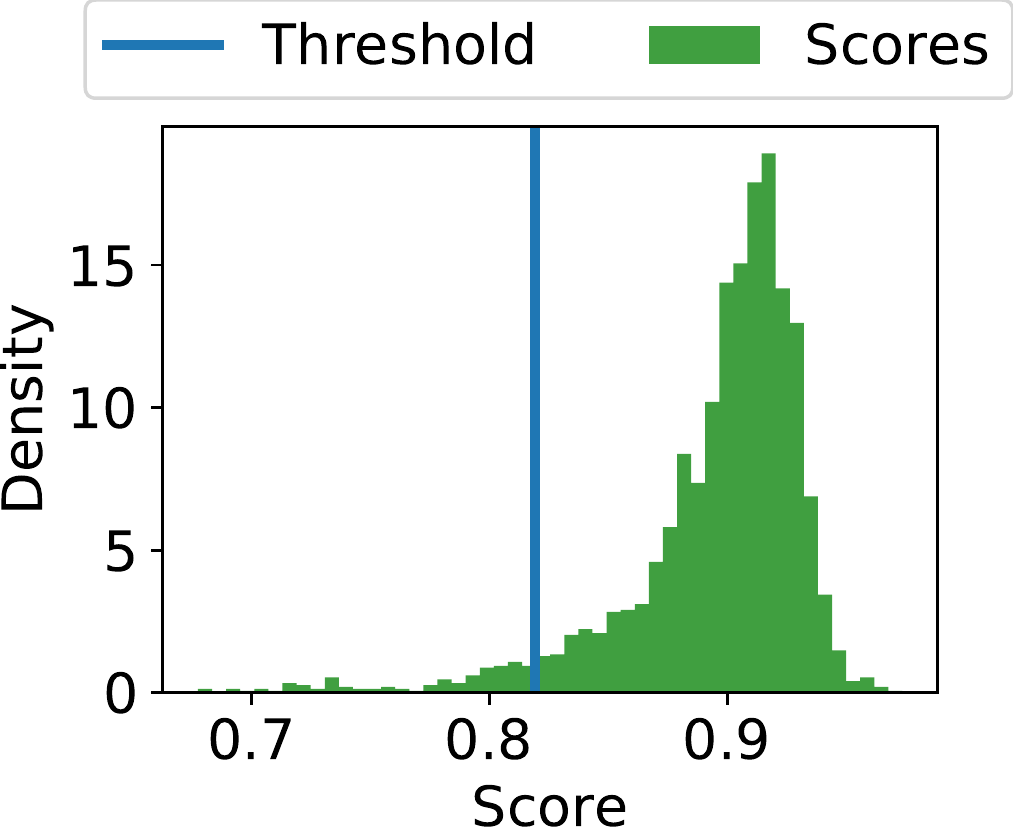}
					\includegraphics[width=0.5\columnwidth]{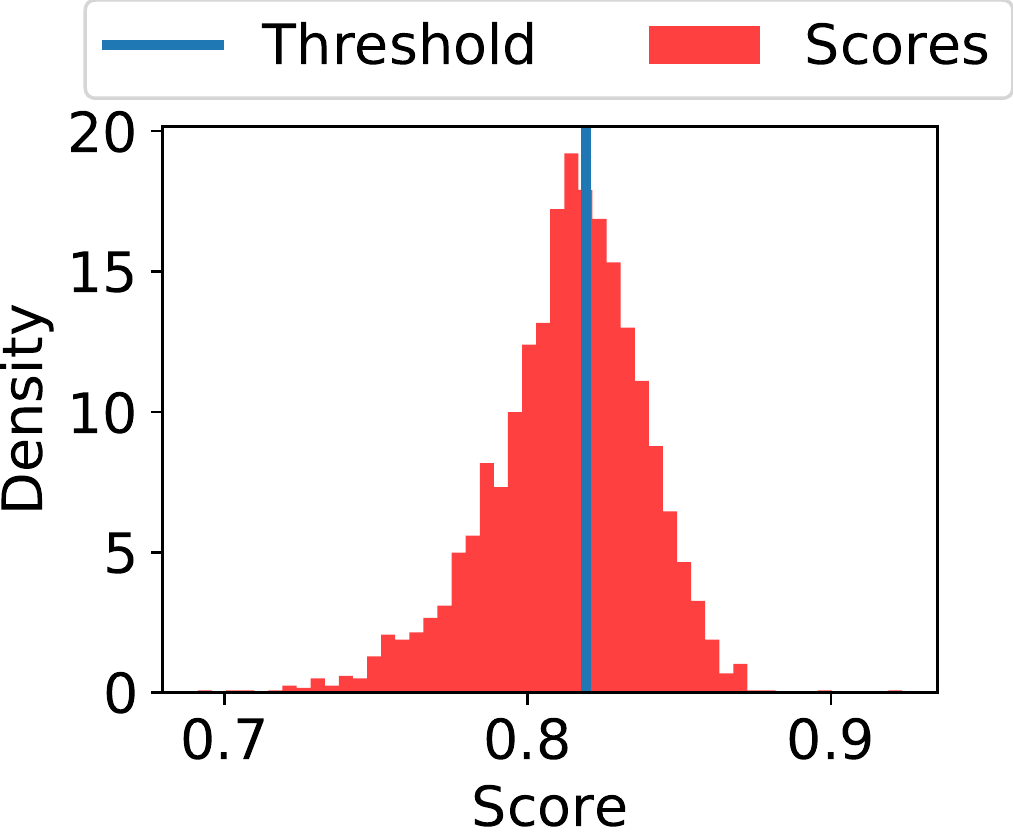}}
		\caption{Distribution of scores for a pair of subjects. Results after 50 stochastic forward passes. On the right a correct prediction with low entropy of $0.27$ is shown. On the left a false prediction with high entropy of $0.98$ is shown.}
		\label{fig:morphing_correct_false}
	\end{center}
\end{figure}

From Figure \ref{fig:entropy_morphing_attack} we can see that the model trained on less data has an overall higher predictive entropy, even if most of the predictions are correct. It also more often fails to discern faces that belong to different subjects, which results in a high false acceptance rate. This behavior expresses a "lack of knowledge", and is often referred to as epistemic or model uncertainty \cite{Gal2016PhD}. Epistemic uncertainty can be reduced by increasing the amount of observed data, which is what we observe with the second model. It was trained using much more examples and has an overall lower uncertainty, which peaks at a blending factor of $0.5$. Moreover, the entropy curve is asymmetric and monotonously rising, even if the accuracy remains unchanged. This might indicate that some subjects are more susceptible to morphing attacks than others (see Fig. \ref{fig:morphing_correct_false}).

\section{Conclusions}
\label{sec:Conclusions}
We propose a novel approach for obtaining uncertainty estimates in deep residual neural networks. We show that our approach produces meaningful uncertainty estimates and can be implemented with minimal effort. We also demonstrate an application of our method to face verification and show its effectiveness in a morphing attack scenario.

Despite its effectiveness, our method has some limitations. The approach can only be applied to architectures with skip connections, such as residual networks or densely connected networks. Another limitation is the fact that shallow networks are more susceptible to stochastic perturbations than deeper ones. Obtaining "fine-grained" uncertainty estimates in shallow networks is more difficult as they have less capacity for redundancy among the residual units. 

In future work, we would like to further evaluate our approach in real-world applications. Furthermore, we would like to study the influence of batch normalization on our method i more detail. Finally, more research has to be done on the optimal drop rates for each residual block. One possibility is to learn this hyper-parameter during the training process.


{\small
\bibliographystyle{ieee}
\bibliography{References}
}

\end{document}